\documentclass[conference]{IEEEtran}
\IEEEoverridecommandlockouts
\usepackage{cite}
\usepackage{amsmath,amssymb,amsfonts}
\usepackage{algorithmic}
\usepackage{graphicx}
\usepackage{amsmath}
\usepackage{animate}
\usepackage{amssymb}
\usepackage{multirow}
\usepackage{subfigure}
\usepackage{color,soul}
\usepackage{textcomp}
\usepackage{xcolor}
\usepackage[breaklinks=true,bookmarks=false]{hyperref}
\usepackage{xcolor}
\hypersetup{
    colorlinks=true,
    linkcolor=teal,
    citecolor=green,
    filecolor=magenta,      
    urlcolor=violet,
}
\usepackage{tablefootnote}

\newcommand{\etal}{\textit{et al}.}
\usepackage{tabularx}
\def\BibTeX{{\rm B\kern-.05em{\sc i\kern-.025em b}\kern-.08em
    T\kern-.1667em\lower.7ex\hbox{E}\kern-.125emX}}
\begin{document}

\title{Spatio-Temporal FAST 3D Convolutions for Human Action Recognition}

% Commented out for blind review.
\author{\IEEEauthorblockN{Alexandros Stergiou}
\IEEEauthorblockA{\textit{Department of Information and Computer Sciences} \\
\textit{Utrecht University}\\
Utrecht, Netherlands \\
a.g.stergiou@uu.nl}
\and
\IEEEauthorblockN{Ronald Poppe}
\IEEEauthorblockA{\textit{Department of Information and Computer Sciences} \\
\textit{Utrecht University}\\
Utrecht, Netherlands \\
r.w.poppe@uu.nl}
}

\maketitle

\begin{abstract}
Effective processing of video input is essential for the recognition of temporally varying events such as human actions. Motivated by the often distinctive temporal characteristics of actions in either horizontal or vertical direction, we introduce a novel convolution block for CNN architectures with video input. Our proposed Fractioned Adjacent Spatial and Temporal (FAST) 3D convolutions are a natural decomposition of a regular 3D convolution. Each convolution block consist of three sequential convolution operations: a 2D spatial convolution followed by spatio-temporal convolutions in the horizontal and vertical direction, respectively. Additionally, we introduce a FAST variant that treats horizontal and vertical motion in parallel. Experiments on benchmark action recognition datasets UCF-101 and HMDB-51 with ResNet architectures demonstrate consistent increased performance of FAST 3D convolution blocks over traditional 3D convolutions. The lower validation loss indicates better generalization, especially for deeper networks. We also evaluate the performance of CNN architectures with similar memory requirements, based either on Two-stream networks or with 3D convolution blocks. DenseNet-121 with FAST 3D convolutions was shown to perform best, giving further evidence of the merits of the decoupled spatio-temporal convolutions.
\end{abstract}


\section{Introduction}
\label{Introduction}
% Action recognition task and brief overview of feature extraction methods
The recognition of human actions in videos remains a challenging task. The current state-of-the-art is obtained using approaches based on convolutional neural networks (CNNs). A large number of increasingly complex network architectures have been introduced to deal with the complexity and variation of the visual performance of human actions in videos \cite{herath2017going,stergiou2018understanding}.

% basis remains dealing with the variation in spatial and temporal domain
Independent from the network architecture, each CNN needs to process the visual input effectively. For static images, 2D convolutions focus on salient spatial patterns. But human actions are characterized by movement over time. To this end, researchers have turned their attention to the spatio-temporal modeling of actions from video. One line of approach that started with Two-stream networks~\cite{simonyan2014two} considers movement modeled as optical flow. The spatial (image) and temporal (optical flow) input is processed independently, which hinders the modeling of characteristic spatio-temporal patterns such as an upwards moving arm.

\begin{figure}
\subfigure[XY]{\includegraphics[width=0.32\columnwidth]{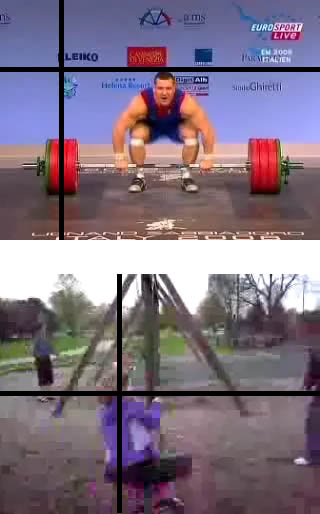}}~
\subfigure[XT]{\includegraphics[width=0.32\columnwidth]{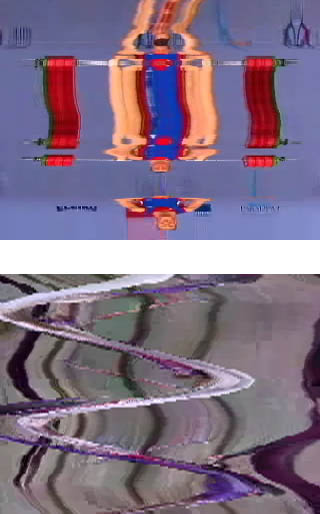}}~
\subfigure[YT]{\includegraphics[width=0.32\columnwidth]{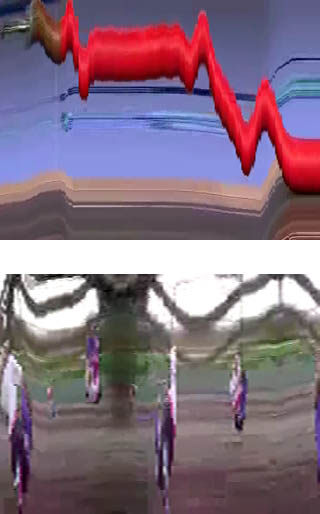}}
\caption{\textbf{Characteristic spatio-temporal motion patterns} Example sequences with motion patterns in the vertical (YT) and horizontal (XT) spatio-temporal domains. Black lines in the first frame indicate where the slices have been made over time.}
\label{fig:movements}
\end{figure}

% 3D convolutions and the role of training data
Alternatively, 3D convolutions~\cite{ji20133d} operate on the spatial and temporal dimensions jointly. This enables the modeling of specific spatio-temporal patterns such as a change in direction of the hand when waving. The modeling of these characteristics requires a large number of parameters in each 3D convolution filter. These parameters are estimated from relevant video data. In contrast to the high number of data samples in image datasets such as ImageNet~\cite{deng2009imagenet}, video-based datasets for specific tasks such as action recognition are composed of significantly less data. UCF-101~\cite{soomro2012ucf101} and HMDB-51~\cite{kuehne2011hmdb} are common benchmarks for video action recognition models, but only include a moderate number of classes and examples. The recent ActivityNet \cite{heilbron2015activitynet} and Kinetics \cite{kay2017kinetics} datasets contain more data, but the number of available examples of each action remains limited, typically in the order of 200-600 videos per class.

\begin{figure}
\begin{center}
\includegraphics[width=.8 \columnwidth]{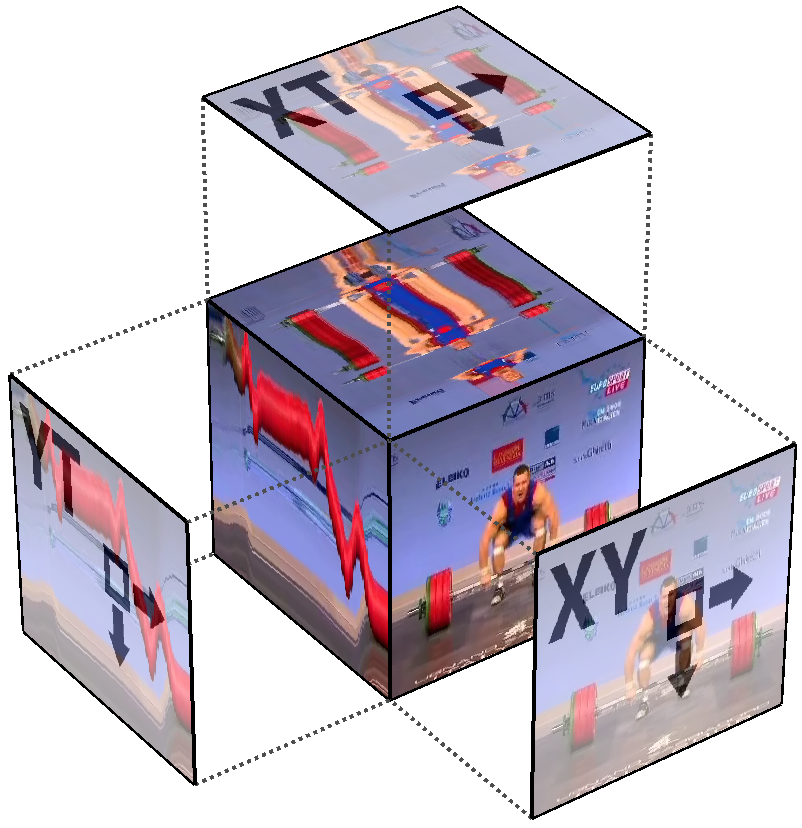}
\caption{\textbf{FAST 3D convolution} Example of a cubic input volume, with the spatial convolution in the XY plane, and two spatio-temporal convolutions in the XT and YT planes.}
\label{fig:spatiotemp_volume}
\end{center}
\end{figure}

% our contribution
On a large number of different CNN architectures and datasets, 3D convolutions generally outperform 2D convolutions by a clear margin. This comes at a cost of estimating more parameters for 3D kernels. This increases the risk of overfitting. This is especially true because of the specificity of the patterns that can be modeled with a 3D convolution. Therefore, we argue that spatio-temporal convolutions of video data with fewer parameters is beneficial for the current availability of training data per action class. We observe that the motion in most actions is relatively simple, and can be described well in terms of horizontal and vertical movement (see Figure~\ref{fig:movements}). We therefore propose Fractioned Adjacent Spatial and Temporal (FAST) 3D convolutions, a novel 3D convolution block. Specifically, we divide 3D convolution kernels into a single spatial 2D kernel responsible for discovering descriptive appearance features and two additional, orthogonal, spatio-temporal kernels that focus on distinctive motion patterns in horizontal and vertical direction respectively.

% advantages
FAST 3D convolutions have several advantages. By splitting the filters, the number of parameters for each convolution operation is reduced so we can construct deeper networks with similar memory requirements as with regular 3D convolutions. Also, the number of non-linearities in the model is triple the amount of that of the original 3D convolutions. Both these advantages allow us to model complex spatio-temporal patterns. We demonstrate improved performance of FAST 3D convolutions over regular 3D convolutions. Finally, the decomposition of 3D convolutional operations decreases overfitting. We show on benchmark datasets that validation loss is reduced, demonstrating that features are learned more efficiently by the model.

% outline
In the next section, we discuss related work on action recognition from video. We then introduce the FAST 3D convolution blocks in Section~\ref{FASTConvs}. Our evaluations on benchmark datasets with various network structures appears in Section~\ref{Experimental Results}. We conclude with promising directions of further research.

%------------------------------------------------------------------------
\section{Related work}
\label{Related Work}
% Video recognition hand-crafted methods
Initial progress in human action recognition has been achieved using low-level handcrafted features including Histograms of Oriented Gradients (HOG), Histograms of Oriented Flow (HOF), Motion Boundary Histograms (MBH)~\cite{wang2013dense} and SIFT \cite{lowe1999object}. Feature representations at the frame or sequence level were aggregated into bag-of-words or Fisher vector representations \cite{gao2016constrained, oneata2013action} and classified as action classes. To deal with correlations between low-level image features in space, mid-level representations such as Poselets~\cite{bourdev2010detecting} and Deformable Part Models (DPM, \cite{felzenszwalb2010object}) have been introduced. These representations focused on the shape and movement of the human body.

% handcrafted -> cnn
While these methods have seen an increased sophistication and performance on benchmark datasets, their handcrafted nature leaves room for improvement. In contrast, Convolutional Neural Networks (CNNs) perform feature extraction using convolutional filters in a hierarchical fashion. This provides more flexibility and allows for the extraction of a large range of low- and mid-level patterns. The use of CNNs has been extended to video by considering sequences of frames as input \cite{karpathy2014large}. This approach allows for the modeling of temporal patterns typical for human actions.

% two-stream networks
An alternative approach to model temporal characteristics is to use optical flow as input in addition to images. Two-stream networks \cite{simonyan2014two} provided the basis for other works including Temporal Segment Networks (TSN, \cite{wang2016temporal}), Temporal Linear Encoding (TLE, \cite{diba2017deep}) and spatio-temporal Regional CNNs \cite{mavroudi2017deep, peng2016multi, saha2016deep, wang2016two}. While these works can model spatio-temporal patterns in videos, optical flow might not be the most effective and efficient way of dealing with the temporal nature of actions. Moreover, the two sources of input are largely processed independently. 

% 3D convolutions
Another approach to extract spatio-temporal patterns is to extend 2D image convolutions to 3D video convolutions \cite{baccouche2011sequential,ji20133d}. Tran \etal ~\cite{tran2015learning} were the first to demonstrate this approach in a deep architecture. Others have also proposed a combination of 3D convolutions and 2D convolutions in order give more weight to the spatial aspect of action recognition \cite{zhou2018mict}. More recently, Carreira and Zisserman~\cite{carreira2017quo} have achieved state-of-the-art performance by combining 3D convolutions in a two-stream network. They pre-trained their I3D network using increasingly complex data. In a last step, they use the comprehensive Kinetics dataset to fine-tune the parameters of their network.

% drawbacks of 3D convolutions and decomposing 2D convolutions
The number of parameters required for each 3D convolutional filter is relatively large. Deep architectures that use 3D convolution blocks tend to overfit because too many parameters need to be estimated from a limited amount of training data. Even with the increasing availability of training data, a decomposition of the 3D convolution filters decreases the risk of overfitting. A similar observation was made for 2D, where convolution kernels that considered the three color channels where split into sequential spatial and color convolutions \cite{kaiser2018depthwise}. This not only reduces the number of parameters to train, but also reduces the discrepancy between training and validation loss.

% decomposing 3D convolutions
For 3D, there have been several approaches to decompose the convolution filters. Tran \etal~\cite{tran2018closer} decomposed ($t \times w \times h$, with $t$ the number of frames and $w$ and $h$ the width and height of the filter, respectively) 3D convolutions into spatial ($1 \times w \times h$) and temporal ($t \times 1 \times 1$) filters. The temporal filters model the variation of pixel values over time. Qiu \etal \cite{qiu2017learning} experimented with this and other decomposed convolution blocks, and added residual connections.

\begin{figure*}
\begin{center}
\subfigure[3D]{\includegraphics[height=6cm,trim=0.15in 4.2in 10.2in 0.2in,clip]{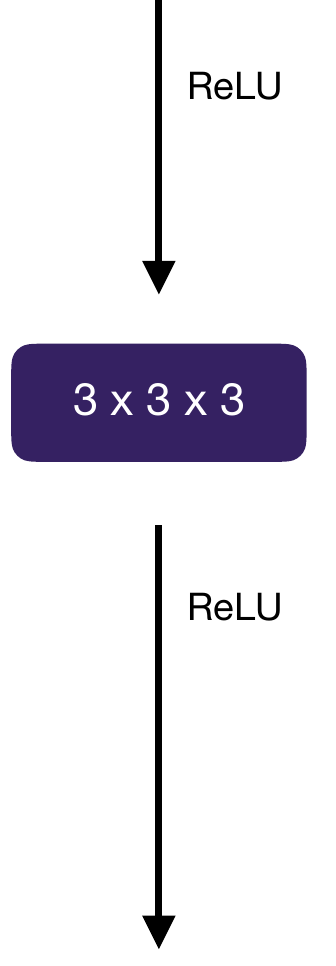}}~
\subfigure[(2+1)D]{\includegraphics[height=6cm,trim=1.7in 4.2in 8.3in 0.2in,clip]{Convolution_blocks.pdf}}~
\subfigure[P3D-B]{\includegraphics[height=6cm,trim=3.6in 4.2in 5.1in 0.2in,clip]{Convolution_blocks.pdf}}~
\subfigure[FAST]{\includegraphics[height=6cm,trim=6.7in 4.2in 3.4in 0.2in,clip]{Convolution_blocks.pdf}}~
\subfigure[split-FAST]{\includegraphics[height=6cm,trim=8.55in 4.2in 0.35in 0.2in,clip]{Convolution_blocks.pdf}}
\end{center}
   \caption{\textbf{Convolution blocks} Design overviews of various convolution blocks, color coded for convolution type: 3D (purple), spatial 2D (magenta), 1D temporal (dark blue) and 2D spatio-temporal (light blue). The depth of the input (i.e., the number of color channels) is omitted for clarity.}
\label{fig:spatial_temporal_convs}
\end{figure*}

% motivation
We argue that purely temporal filters will not be able to model motion boundaries. Our work is motivated by the observation that many motions have distinctive characteristics in horizontal and vertical directions (see Figure~\ref{fig:movements} for an example). We therefore propose Fractioned Adjacent Spatial and Temporal (FAST) 3D convolutions, a decomposition of 3D convolutions into a spatial, and horizontal motion and a vertical motion part. For the motion parts, we use two orthogonal 2D convolutions that essentially treat a local part of a video as $XT$ (horizontal) and $YT$ (vertical) slices, reminiscent of early work by Niyogi \etal \cite{niyogi1994analyzing}.

% contribution
We evaluate the performance of the novel FAST convolution block in a number of popular architectures and on well-known action recognition benchmark datasets UCF-101 and HMDB-51. While there is room for improvement in the absolute performance by using more sophisticated network structures (e.g., \cite{chollet2017xception,he2016deep,hu2018squeeze,huang2017densely,szegedy2016rethinking}), pre-training on larger datasets (e.g. \cite{carreira2017quo}) and increasing batch sizes (e.g. \cite{chen2018multifiber}), we consistently report increased classification performance, while at the same time observe lower validation losses. This suggests that we can effectively and efficiently learn characteristic spatio-temporal patterns, with increasing generalization abilities of the trained networks.

%-------------------------------------------------------------------------
\section{FAST 3D convolution blocks} \label{Implementation}
In this section we briefly describe 3D convolutions and recently introduced variations. We then provide a detailed description of the proposed FAST 3D convolutions. We also introduce an alternative split-FAST block in which the temporal convolutions are performed in separate pathways.

\subsection{3D convolutions and variants} %3D, (2+1)D and Pseudo Convolutions}
\label{3DConvolutions}
% 3D Convs
We consider an input video, denoted as a spatio-temporal volume $\mathbf{X}$ with size $F \times H \times W \times D$, made up of a temporal dimension with $F$ frames and two spatial dimensions of height $H$ and width $W$ in pixels. $D$ corresponds to the depth of each pixel, typically the network input's number color channels or activation maps in intermediate layers. At each layer $i$ of the network, the input volume $\mathbf{X_i}$ is processed with $n_i$ kernels. For clarity of presentation, we omit the indexing on the layer $i$ and the kernel $j$, $1 \leq j \leq n_i$. Each kernel $K$ is a four-dimensional tensor $K \in \mathbb{R}^{f \times h \times w \times d}$, with $f$, $h$, $w$ and $d$ the size of the kernel in the temporal, horizontal and vertical spatial and depth dimension, respectively. For a 3D convolution of video volume $\mathbf{X}$ with kernel $K$, activation map $Y$ becomes:

\begin{equation}
Y = K \otimes X
\label{eq:3dconv}
\end{equation}

% (2+1)D Convs
\paragraph{(2+1)D convolutions} Instead of processing an input video with a four-dimensional kernel, Tran \etal~\cite{tran2018closer} separate the kernel into a three-dimensional spatial kernel $K_s \in \mathbb{R}^{1 \times h \times w \times d}$ and a temporal tensor $K_t \in \mathbb{R}^{f \times 1 \times 1 \times d}$. $K_s$ essentially operates as a 2D convolution, whereas $K_t$ looks at the change in pixel intensity over time. See Figure~\ref{fig:spatial_temporal_convs}(b). Convolutions with the two kernels is performed subsequently:

\begin{equation}
Y = K_t \otimes (K_s \otimes X)
\label{eq:2+1d}
\end{equation}

% characteristics
This method, termed (2+1)D, doubles the number of non-linearities and can therefore increase the complexity of the feature mapping. The decoupling of the four-dimensional kernel for 3D convolutions into a three-dimensional and a one-dimensional kernel results in an overall lower number of parameters that need to be estimated and updated for each convolution block. The memory required for activation maps during training is somewhat larger than that of normal 3D convolutions because the number of activation maps stored and updated by the system per layer doubles.

% pseudo
\paragraph{Pseudo convolutions} Variations of 3D convolutions have also included the use of different configurations in terms of order and connections of the convolutions. These convolution blocks have been termed Pseudo Convolutions, or P3D~\cite{qiu2017learning}. Specifically, three different variants have been introduced. P3D-A is similar to (2+1)D convolutions, but with a skip connection. Instead of processing the spatial and temporal convolutions in sequence, P3D-B processes them in parallel, see Figure~\ref{fig:spatial_temporal_convs}(c). The outputs are then accumulated:

\begin{equation}
Y = (K_t \otimes X) \oplus (K_s \otimes X)
\label{eq:p3d-b}
\end{equation}

The main advantage of P3D-B over P3D-A is the structural diversity of the architecture. This corresponds to the network focusing on the most informative dimension in each level of the feature extractor. Finally, P3D-C includes two residual connections: one for the spatial convolution block and one for the temporal block. P3D-C can be seen as a compromise of P3D-A and P3D-B.

% conceptual: limitations
For both (2+1)D convolutions and Pseudo convolutions, the temporal convolutions look at changes in the pixel intensity only in the first layers of the CNN. For deeper layers, an increasingly large area is taken into account. Still, characteristic spatio-temporal patterns cannot be modeled because the kernel is only one-dimensional.

\subsection{FAST 3D convolutions} \label{FASTConvs}
% motivation
We introduce a novel convolution block: \textit{Fractioned Adjacent Spatial Temporal} (FAST) Convolutions. This block is motivated by the desire to decompose the four-dimensional 3D convolution kernel, but to also maintain the ability to model spatio-temporal patterns explicitly. The decomposition should lead to less overfitting, especially on smaller training datasets. The explicit modeling of motion is of particular interest in human action recognition, where many classes are characterized by distinct spatio-temporal patterns (see Figure~\ref{fig:movements}).

% architecture
The FAST 3D convolutions block is a combination of three convolutions (see Figure~\ref{fig:spatiotemp_volume}), performed sequentially (see Figure~\ref{fig:spatial_temporal_convs}(d)). Essentially, we split motion into a horizontal and a vertical component. In contrast to (2+1)D convolutions, we do not only model pixel changes over time but consider the temporal dimension jointly with either the horizontal or vertical dimension. We thus apply the convolutions in the XT plane and YT plane of a video volume. This allows us to capitalize on specific horizontal and vertical movements, performed within a limited spatial context. The two spatio-temporal convolution operations are complementary to each other, as they decompose movement in a vertical and horizontal fashion. More complex motions can be modeled by both operations jointly.

% processing
We denote the kernels for the horizontal and vertical spatio-temporal convolutions as $K_{XT}$ and $K_{YT}$, respectively. In addition to these two spatio-temporal convolutions XT and YT, we use a spatial kernel $K_{XY}$ which is a regular 2D kernel. This is in line with the spatial kernel used in (2+1)D convolutions. FAST 3D convolutions are thus operationalized as:
\begin{equation}
Y = K_{YT} \otimes \big(K_{XT} \otimes (K_{XY} \otimes X)\big)
\label{eq:fast}
\end{equation}

The frame-level filter (XY) iterates spatially in each frame, extracting visual characteristics of the scene. The horizontal (XT) and vertical (YT) spatio-temporal kernels iterate through time with the frame's width and height as the auxiliary dimension, respectively.

\subsubsection{Temporally decoupled connections} \label{FASTvariants}
Additionally, we introduce a variant of FAST 3D convolutions with indirect connections between the two spatio-temporal convolutions, schematically shown in Figure~\ref{fig:spatial_temporal_convs}(e). Thus, after the convolution with the spatial filter, the horizontal XT and vertical YT convolutions are performed in parallel and the output is then accumulated. We denote this architecture as split-FAST as the sequence of temporal convolutions is split into two pathways. 

Our intuition behind creating two pathways for the temporal operations is that some movements are characterized predominantly by one of the two operations. For example, in a jumping or push-up motion, distinctive patterns are more likely to be found in the vertical direction. In contrast, we expect that the responses of the XT convolution are much less meaningful. For mainly horizontal movements such as walking or running, we expect the opposite effect. By choosing the most fitting temporal convolution, the model progressively learns the type of movement that each input includes as the most fitting kernels are chosen.

%-------------------------------------------------------------------------
\section{Experimental results}
\label{Experimental Results}

In this section, we compare our proposed FAST 3D convolutional blocks to 3D and (2+1)D convolutions in a ResNet-34~\cite{he2016deep} architecture (Section~\ref{sec:Blockcomparisons}). We focus on human action recognition and present results on UCF-101~\cite{soomro2012ucf101}, a publicly available human action recognition dataset. We then investigate how stable the improvements over 3D convolutions are when the network depth increases (Section~\ref{sec:Revisitingarchitectures}). We also present additional results on HMDB-51~\cite{kuehne2011hmdb}, a second well-known human action recognition dataset. Finally, we compare our results to a number of popular, state-of-the-art implementations (Section~\ref{sec:Networkcomparisons}).

% limitations
The main contribution of this paper is the introduction of a novel convolution block for videos. In this evaluation, our focus is on assessing the merits of this block over previously introduced convolution blocks. Our proposed method is general in the sense that it can be used in a wide range of network architectures, as we demonstrate in this section. Importantly, we do not attempt to achieve state-of-the-art performance. Compared to the architectures that we evaluate on, more complex deep networks and sophisticated (pre)training methods have been proposed in literature. We note that these state-of-the-art networks could benefit from our proposed FAST 3D and split-FAST 3D convolution blocks.

\begin{figure*}[thb]
\begin{center}
\subfigure[3D]{\includegraphics[height=2.5cm]{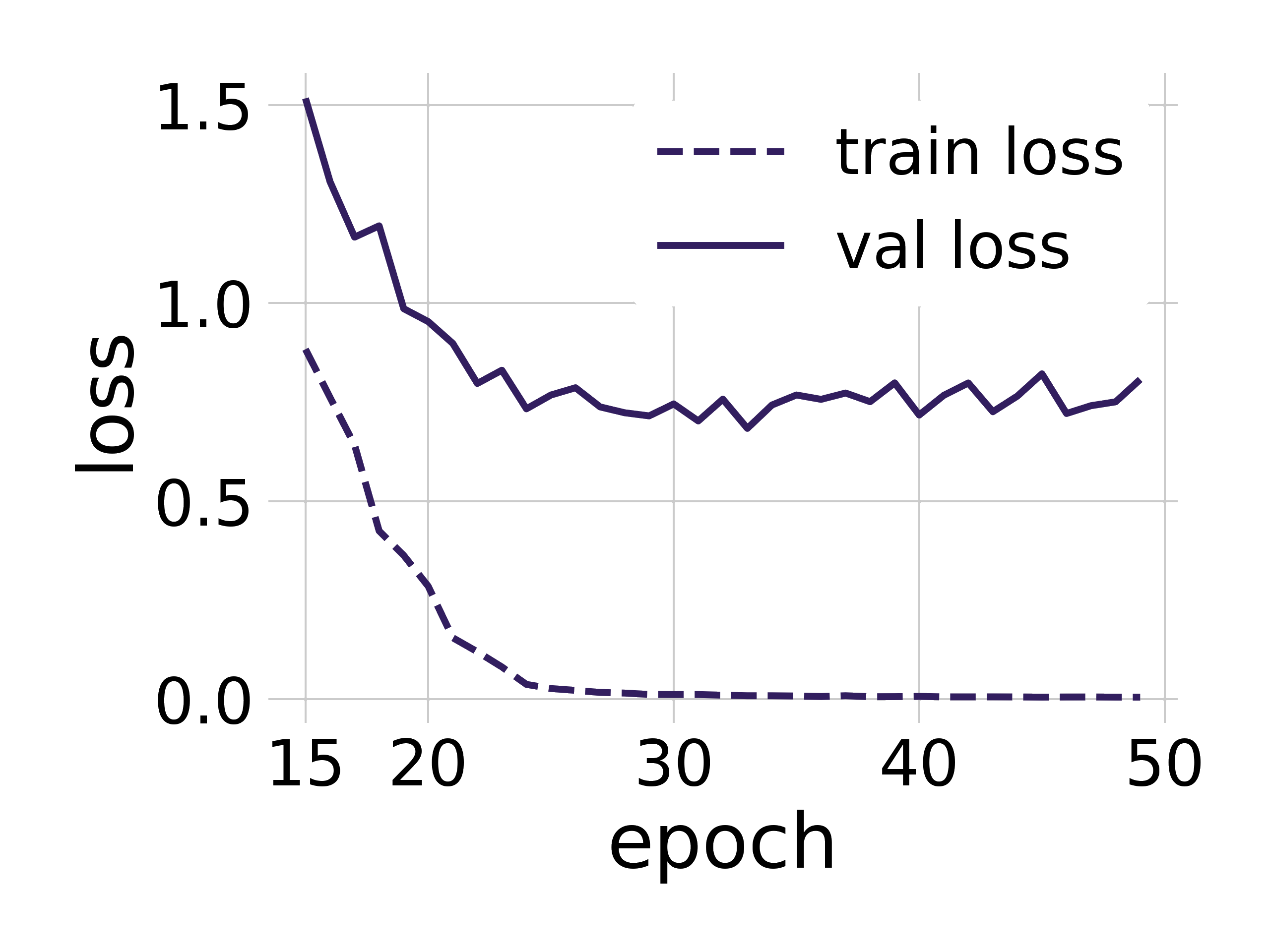}}
\subfigure[(2+1)D]{\includegraphics[height=2.5cm]{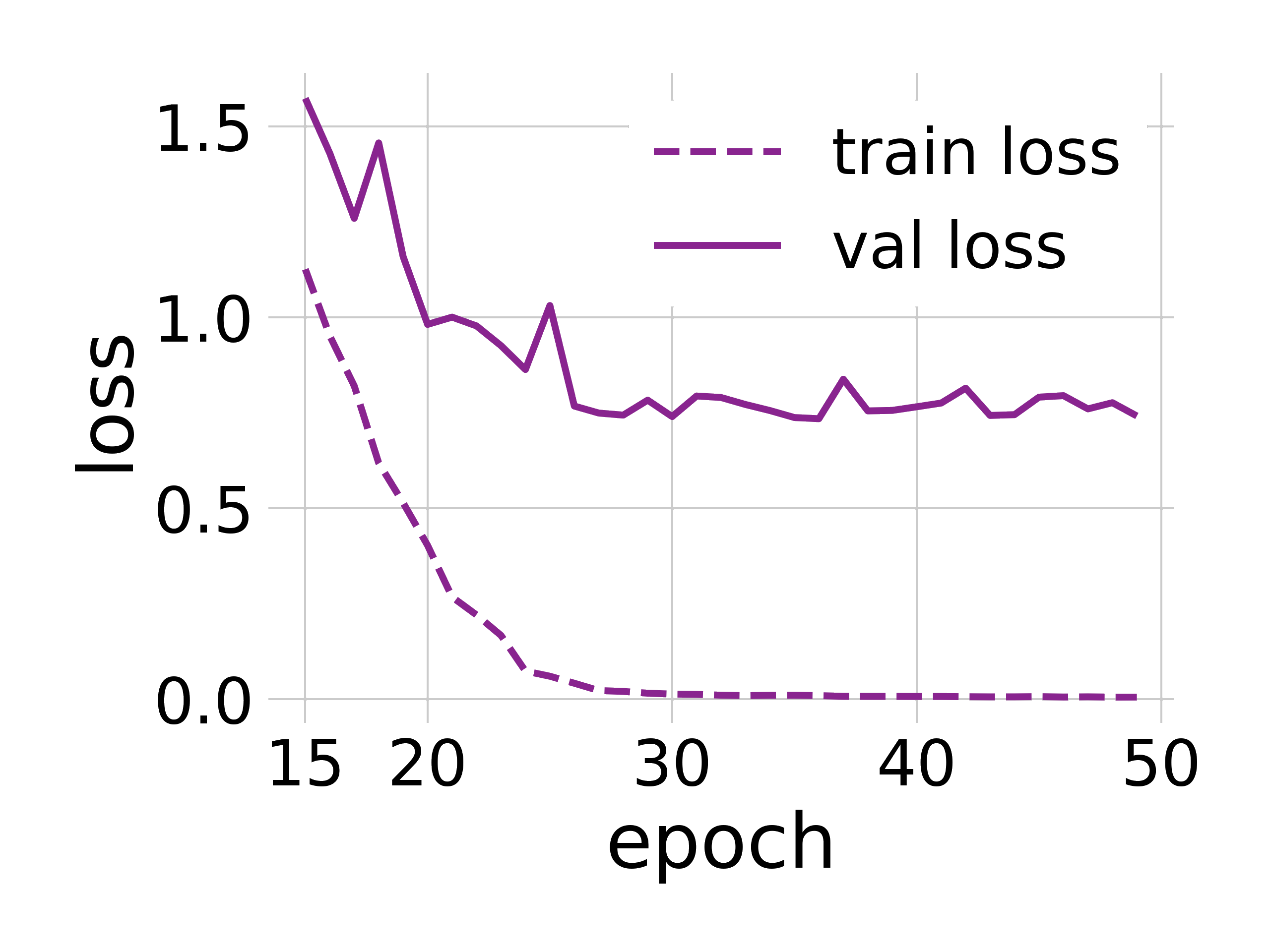}}
\subfigure[FAST]{\includegraphics[height=2.5cm]{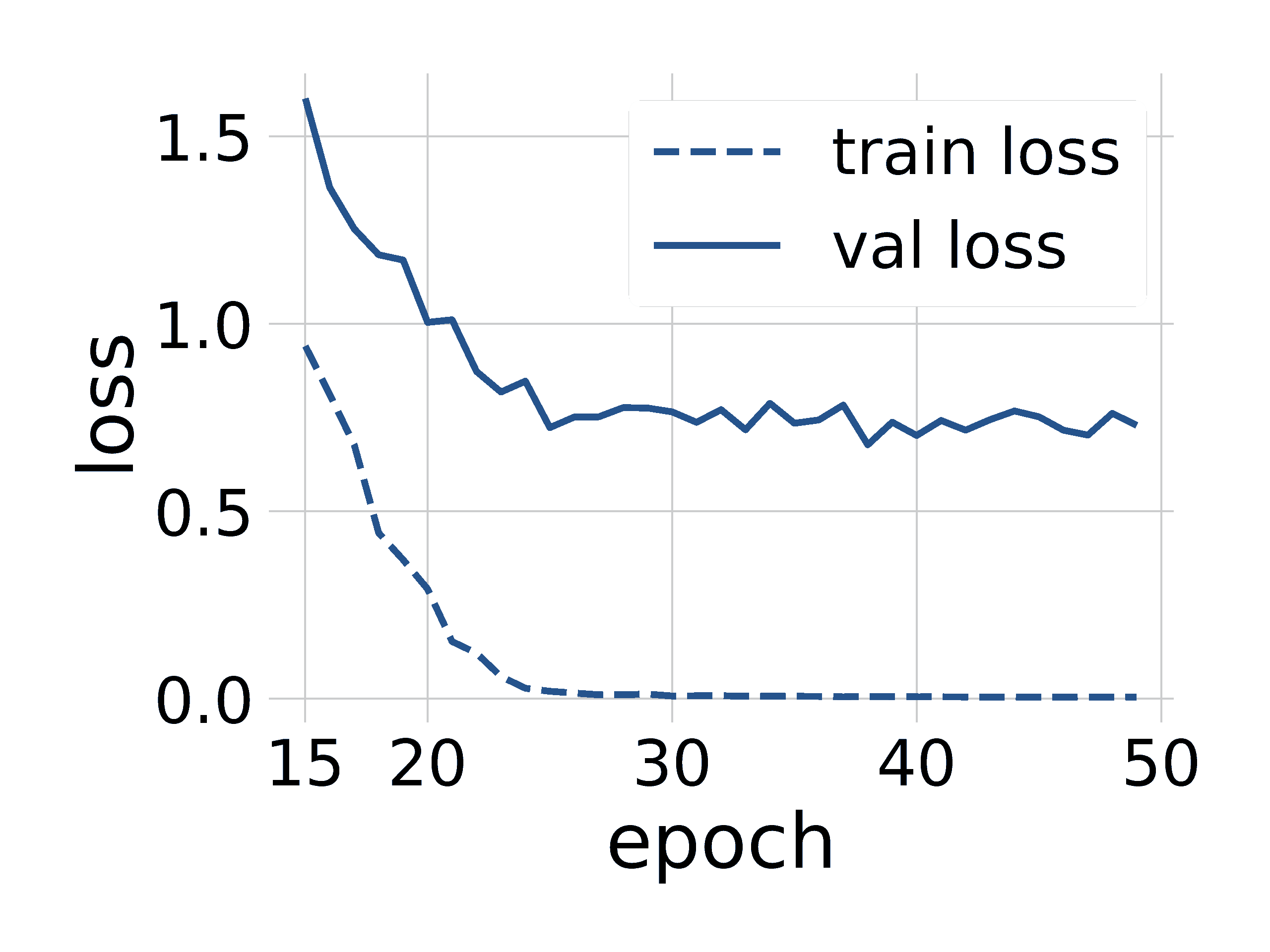}}
\subfigure[split-FAST]{\includegraphics[height=2.5cm]{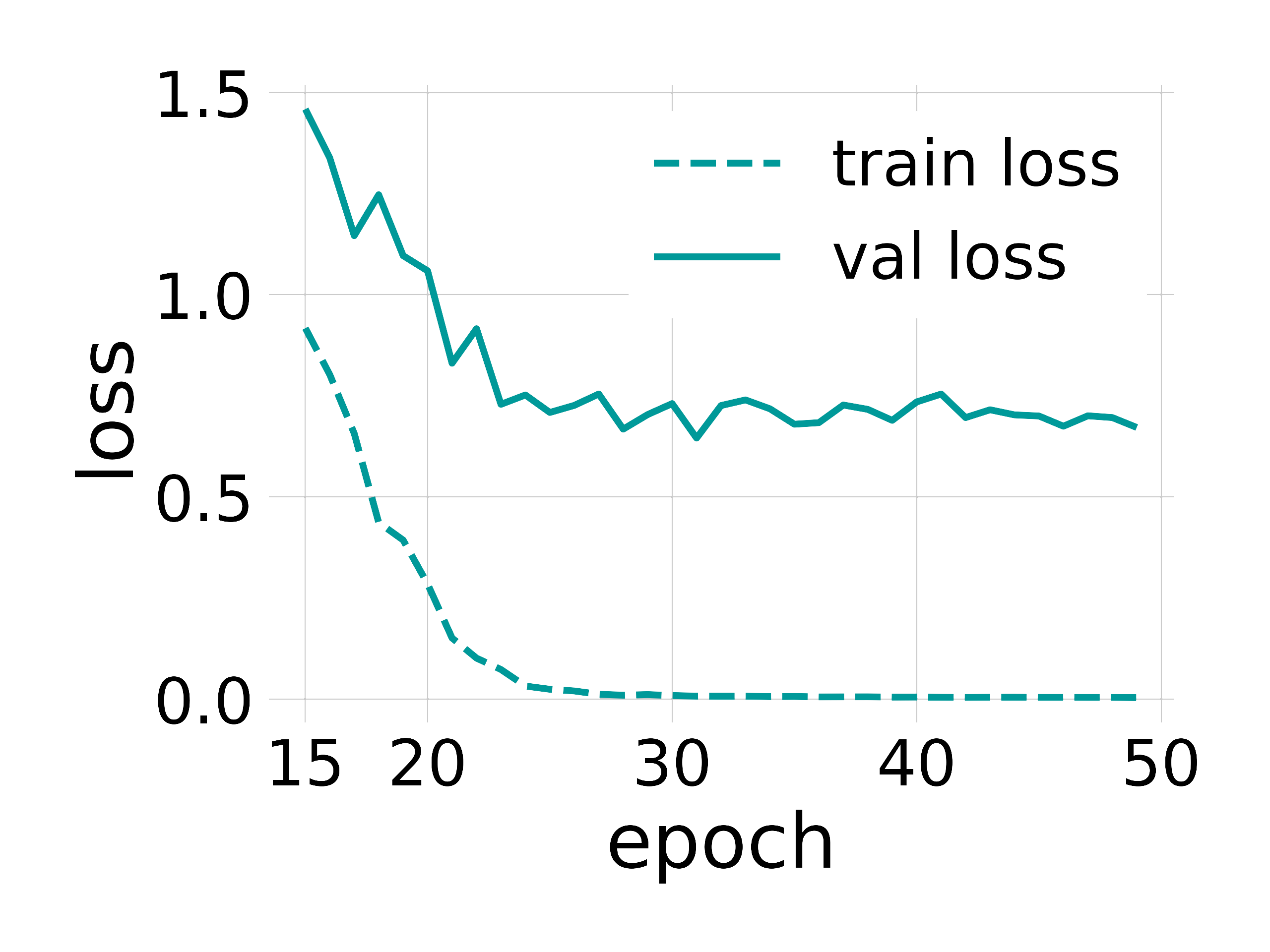}}
\subfigure[Combined]{\includegraphics[height=2.5cm]{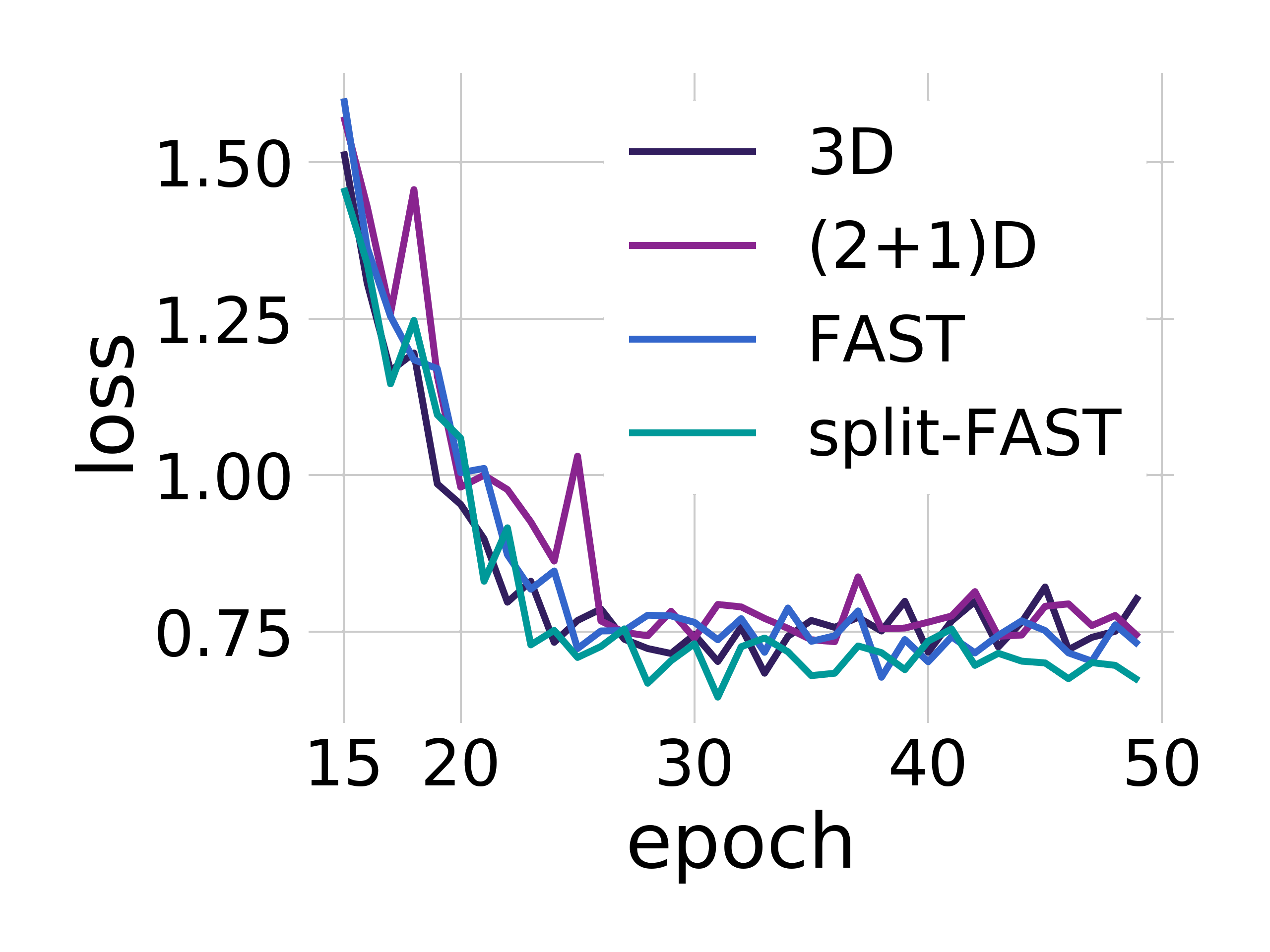}}
\end{center}
   \caption{\textbf{Training and validation loss} for (a) 3D, (b) (2+1)D, (c) FAST and (d) split-FAST, obtained on UCF-101 using ResNet-34. The validation losses for these convolution blocks are combined in (e). The first 15 epochs are omitted to increase the visibility of the final loss values. The classification performances for these models appear in Table~\ref{table:table1}.}
\label{fig:loss_table_1}
\end{figure*}

\subsection{Experiment settings}
\label{Experimentssettings}

\paragraph{Datasets} We evaluate on the UCF-101 and HMDB-51 datasets. Both are standard action recognition benchmark datasets. We have chosen these datasets because they have been widely reported on and their nature and challenges are broadly understood. Compared to ActivityNet and Kinetics, both UCF-101 and HMDB-51`are modest in size, but contain the same type of variations in terms of viewpoint, clutter, action performance and image quality. UCF-101 contains 13,320 videos in 101 classes, HMDB-51 consists of 6,849 clips distributed over 51 classes.

\paragraph{Implementation} For our evaluation of convolution blocks (Section~\ref{sec:Blockcomparisons}), we focus on ResNet models because they are well-understood and provide decent performance for their limited complexity. We also present results for the FAST 3D convolution block in a range of other architectures in Section~\ref{sec:Networkcomparisons}. 

For efficient training, we select 24 frames from each video and resize the frames to a format with a size of $224 \times 224$. The 24 selected frames cover a 2 second window halfway the duration of the sequence. In this volume, we select every second frame. Inputs are normalized to single-float point precision. All experiments are performed with two NVIDIA Tesla P100 GPUs. For the experiments we used a SGD with 0.9 momentum and with a learning rate that uses warm-up restarts \cite{loshchilov2016sgdr} every 5 epochs and ranges between a maximum value of 2e-3 and a minimum of 4e-5, with the maximum value halved at the end of each cycle. We use a Dropout rate of 0.5 for experiments with DenseNet-121. For all implemented methods, we used the parameters reported in the respective papers. The only exception is the batch size, which we set to 8 because of limited available memory. All networks are initialized with weights from ImageNet. For 3D convolutions, we inflated the 2D kernels to 3D, similar to~\cite{carreira2017quo}.

\begin{table}[htb]
\begin{center}
\resizebox{\columnwidth}{!}{%
\begin{tabular}{|l c c c c c|}
\hline
Method & Accuracy & Speed & Depth & Params & GB \\
\hline\hline
3D & 81.14 & 5.40 & 137 & 40.60M & 11.32 \\
\hline\hline
(2+1) D & 81.75 & 4.60 & 147 & 25.84M & 13.35 \\
\hline\hline
FAST & 83.82 & 4.26 & 157 & 43.48M & 12.06 \\
split-FAST & \textbf{85.36} & 4.26 & 157 & 43.48M & 12.06 \\
\hline
FAST (XT only) & 82.88 & 4.88 & 147 & 32.89M & 13.35 \\
FAST (YT only) & 83.18 & 4.83 & 147 & 32.89M & 13.35 \\
\hline
\end{tabular}}
\end{center}
\caption{Comparison between convolution blocks, used in a ResNet-34 on UCF-101. Speed is measured in clips per second and depth corresponds to the total number of layers.\label{table:table1}}
\end{table}

\subsection{Block comparisons} \label{sec:Blockcomparisons}
% 3D block variant comparisons (3D vs. Pseudo vs. (2+1)D vs. FAST)
Table~\ref{table:table1} summarizes the performance of different convolution blocks in a ResNet-34 architecture, trained and evaluated on UCF-101. All networks were trained for the same number of epochs with the same learning rate. Data were fed to the network by four workers so the time required for loading the data was the same.

% accuracy
When comparing the convolution blocks, we see that both FAST variants and (2+1)D convolutions outperform 3D convolutions in terms of accuracy. While (2+1)D convolutions show a modest improvement of 0.61\%, FAST and split-FAST outperform the original 3D convolutions by 2.68\% and 4.22\%, respectively. These results indicate that the modeling of temporal characteristics in terms of changes in pixel values using a $3 \times 1 \times 1$ temporal kernel leaves room for improvement. Clearly, the description of motion in terms of two orthogonal spatio-temporal directions is beneficial for the modeling of human actions. For our given setting, FAST 3D convolutions demonstrate an improvement of 2.07\% over (2+1)D convolutions.

% network depth, parameters and memory
For kernels of size 3 in spatial or temporal dimension, the three 2D kernels used in FAST and split-FAST require slightly more parameters than the single 3D kernel used in 3D convolutions. In comparison, the $3 \times 1 \times 1$ temporal kernel used in (2+1)D convolutions is significantly smaller than 3D and both FAST 3D convolutions. The successive convolutions in both (2+1)D and both FAST 3D convolution blocks increase the depth of the network. Consequently, these blocks have the potential to model more non-linearities in their mapping. As a drawback, the successive convolution operations also increase the memory required to store intermediate activation maps. In addition, in terms of training speed, both (2+1)D and FAST convolutions require more time to learn per batch as the networks are deeper and more updates per pass are required.

% losses
In Table~\ref{fig:loss_table_1}, we show the training and validation losses for 3D, (2+1)D and the two FAST 3D convolutions. While differences in validation loss between the convolution blocks are modest, FAST and split-FAST consistently have lower final losses. This is an indication of less overfitting. It is also an important contribution given the larger number of parameters in the FAST 3D convolution blocks. Apparently, the motion patterns that are modeled in the spatio-temporal XT and YT convolutions are meaningful and generalize to unseen data.

\begin{figure}[htb]
\begin{center}
\subfigure[Validation loss]{\includegraphics[height=3cm]{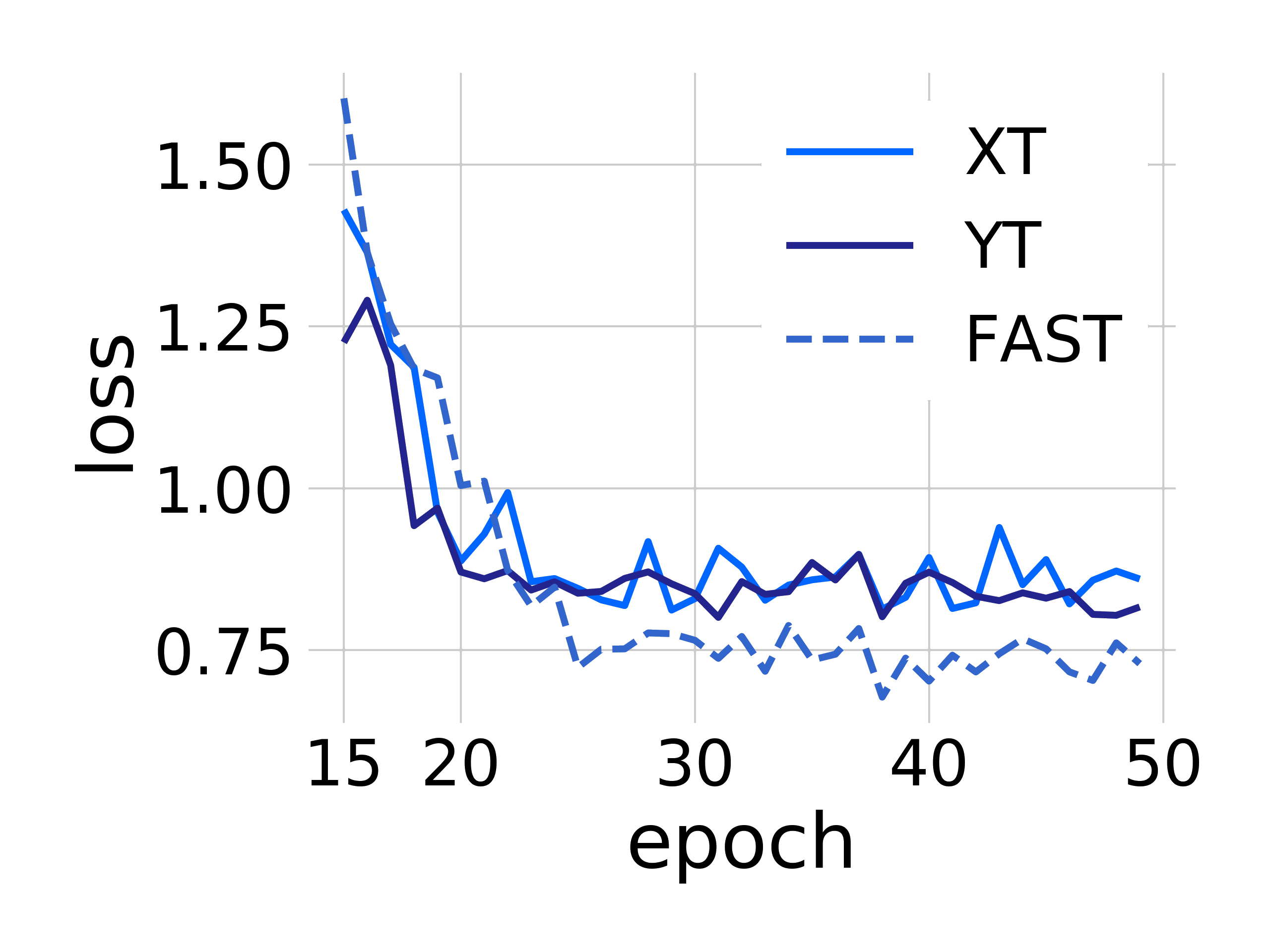}}
\subfigure[Accuracy]{\includegraphics[height=3cm]{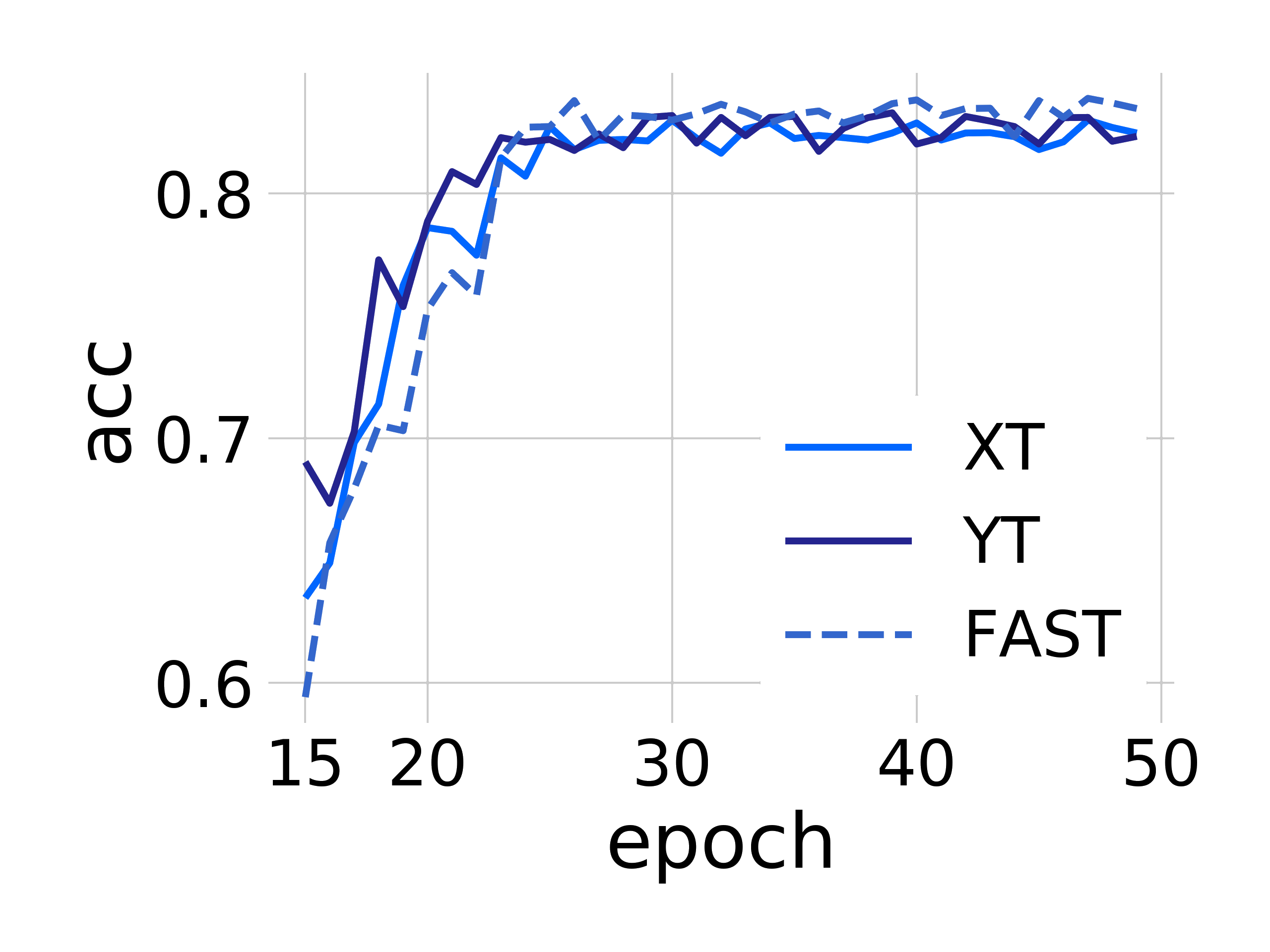}}
\end{center}
   \caption{\textbf{Validation loss and accuracy} for a Resnet-34 on UCF-101. The first 15 epochs are omitted.}
\label{fig:loss_xt_yt}
\end{figure}

% XT and YT slicing
\paragraph{XT and YT convolutions in FAST} In order to analyze the importance of the two spatio-temporal kernels in FAST 3D convolutions, we also evaluated the performance when one of the kernels was omitted. To this end, we replaced one of the 2D spatio-temporal kernels with the $3 \times 1 \times 1$ temporal kernel used in (2+1)D convolutions, but applied in the dimension that we omitted. Specifically, in the XT-only model, we used changes in pixel values along the vertical spatial dimension. For the YT-only model, we only considered pixel changes in the horizontal dimension.

% results
From Table~\ref{table:table1}, it becomes clear that both blocks with an omitted 2D kernel produce sub-par results compared to FAST 3D convolutions. This indicates that important spatio-patterns are missing if one of the two dimensions is not considered (see Figure~\ref{fig:loss_xt_yt}). Since we can model at least spatio-temporal patterns in one direction, both models outperform (2+1)D convolutions, by 1.13\% and 1.43\% for Fast 3D with XT-only and YT-only spatio-temporal kernels, respectively. Again, this demonstrates that characteristic motion patterns are ignored when simply looking at changes in pixel values over time.

% xt vs yt
It appears that vertical motion is more important than horizontal motion. This might be because many of the videos in UCF-101 contain predominantly horizontal panning motion to keep the subject of interest in the center of the view. This might cause horizontal movement to be more related to the camera movement, rather than with specific actions. Still, even individual dimensions contribute to the improvement of the model. The fact that both temporal features together lead to the highest score, is an indication that there is partially complementary information in both directions.

% FAST splits
\paragraph{FAST and split-FAST} The decoupled split-FAST 3D convolution block outperforms 3D convolutions by 4.22\% and FAST 3D convolutions with three sequential convolution operations by 1.54\%. This higher performance suggests that the temporal kernels do not only learn vertical and horizontal movement explicitly, but also more complex movements such as those shown in Figure~\ref{fig:movements}. Since the split-FAST approach further groups actions of small clip segments based on their overall movement across frames, it can more efficiently interpret the type of movement from the vertical and horizontal separation. From Figure~\ref{fig:loss_table_1}, it follows that decoupling the two spatio-temporal convolution operations improves the generalization capabilities of the architecture, judging from the slightly smaller divergence between training and validation losses.

\begin{figure}[htb]
\begin{center}
\subfigure[Validation loss - UCF-101]{\includegraphics[height=3cm]{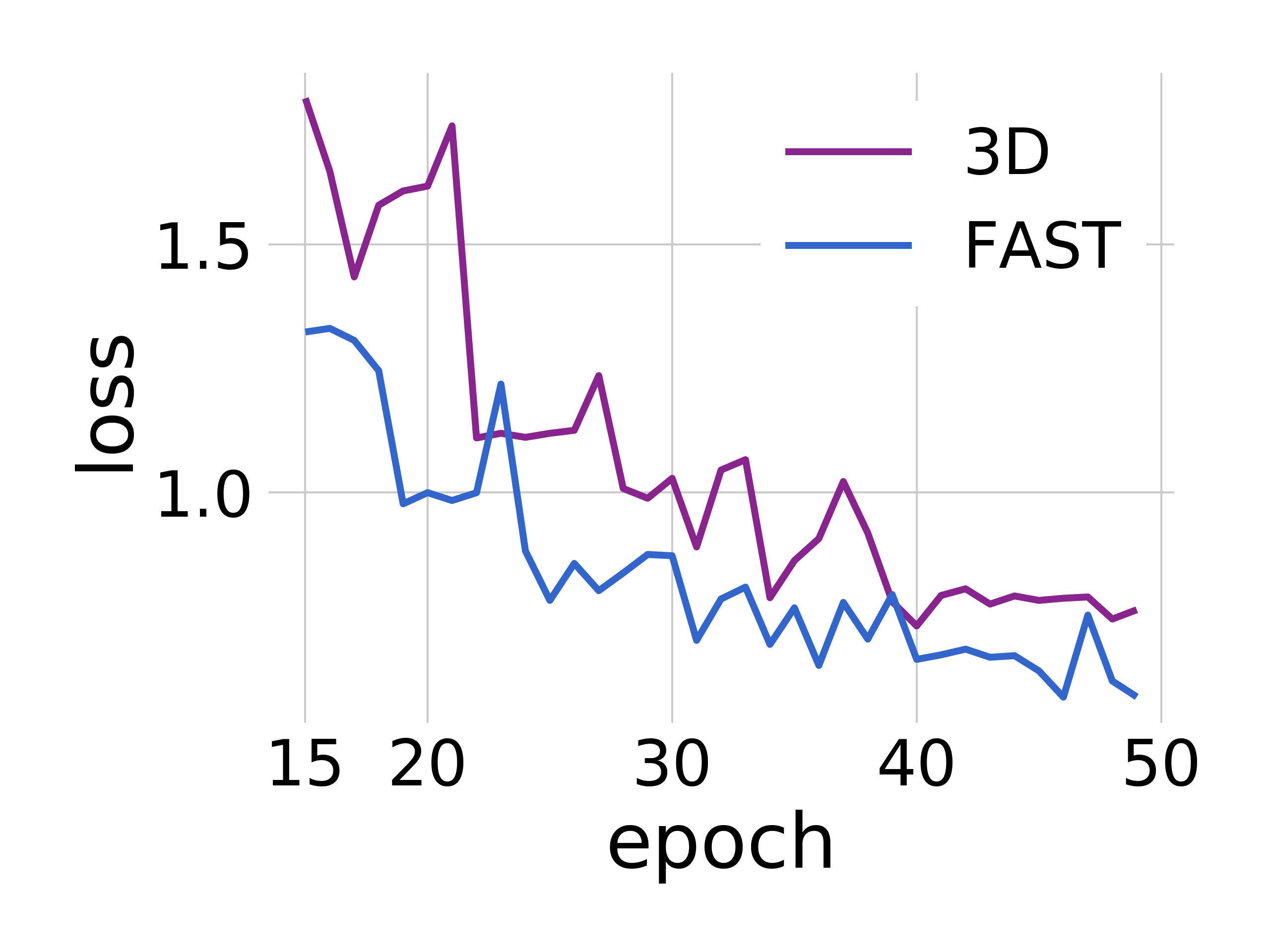}}
\subfigure[Accuracy - UCF-101]{\includegraphics[height=3cm]{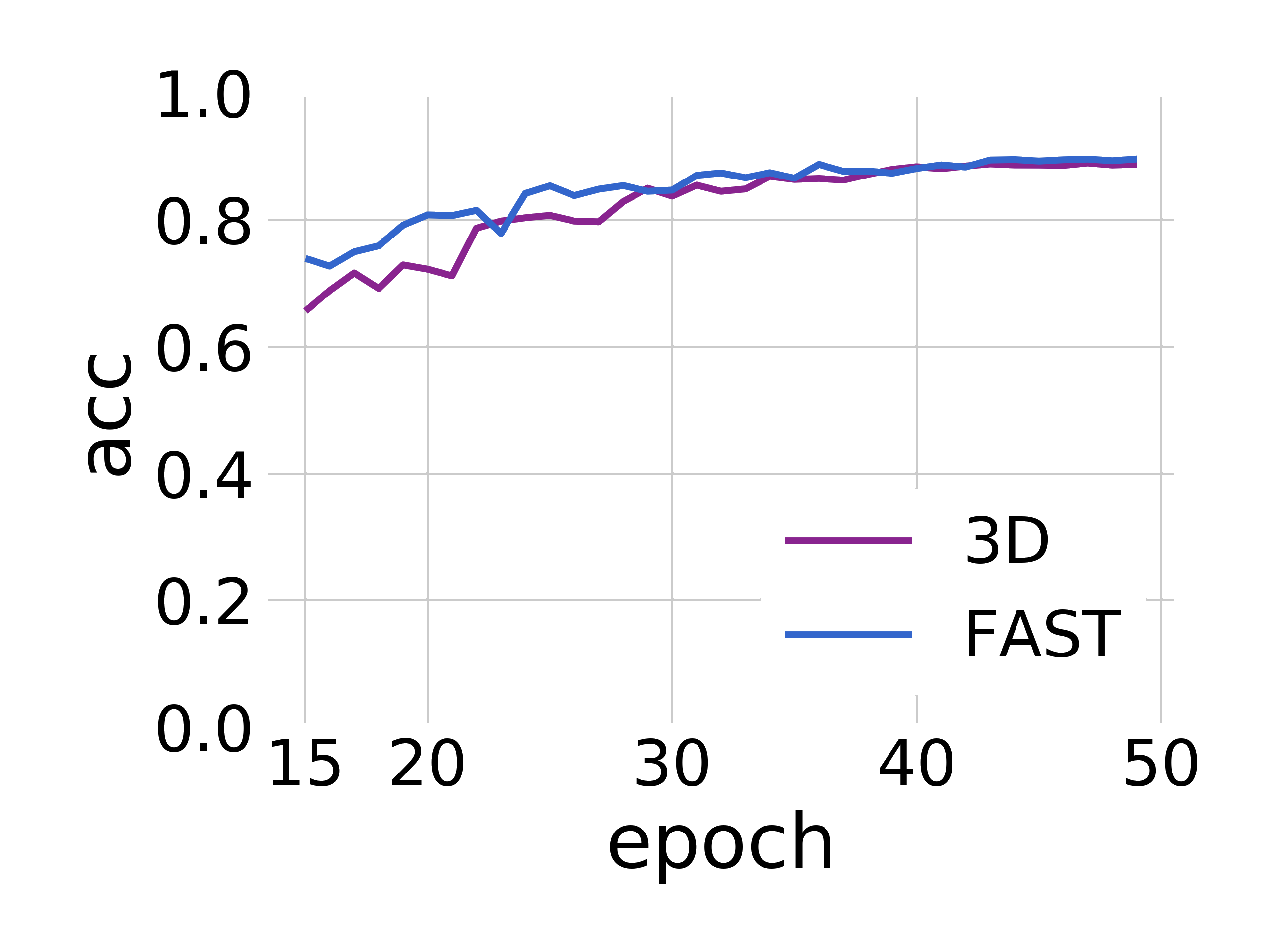}}
\subfigure[Validation loss - HMDB-51]{\includegraphics[height=3cm]{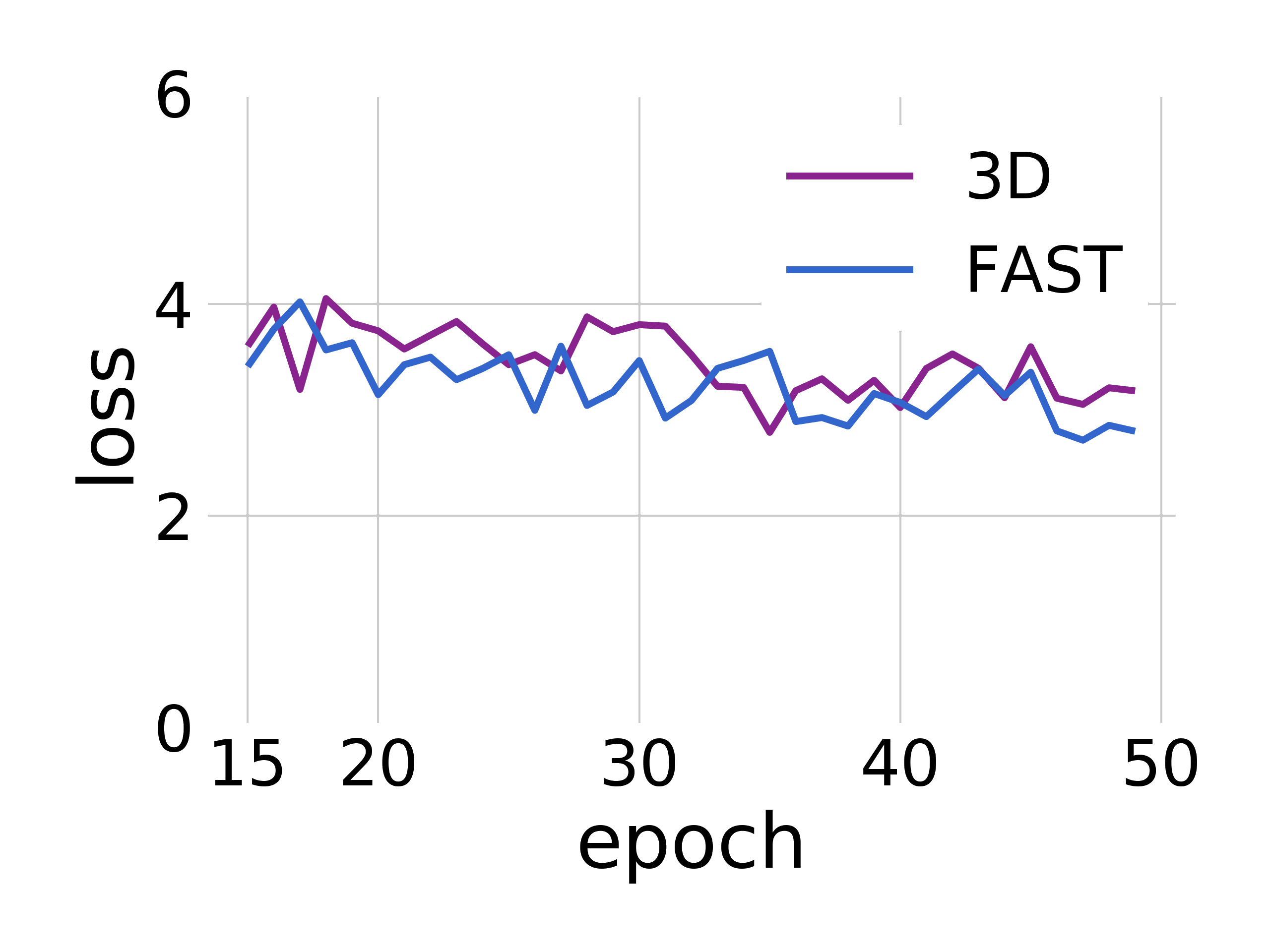}}
\subfigure[Accuracy - HMDB-51]{\includegraphics[height=3cm]{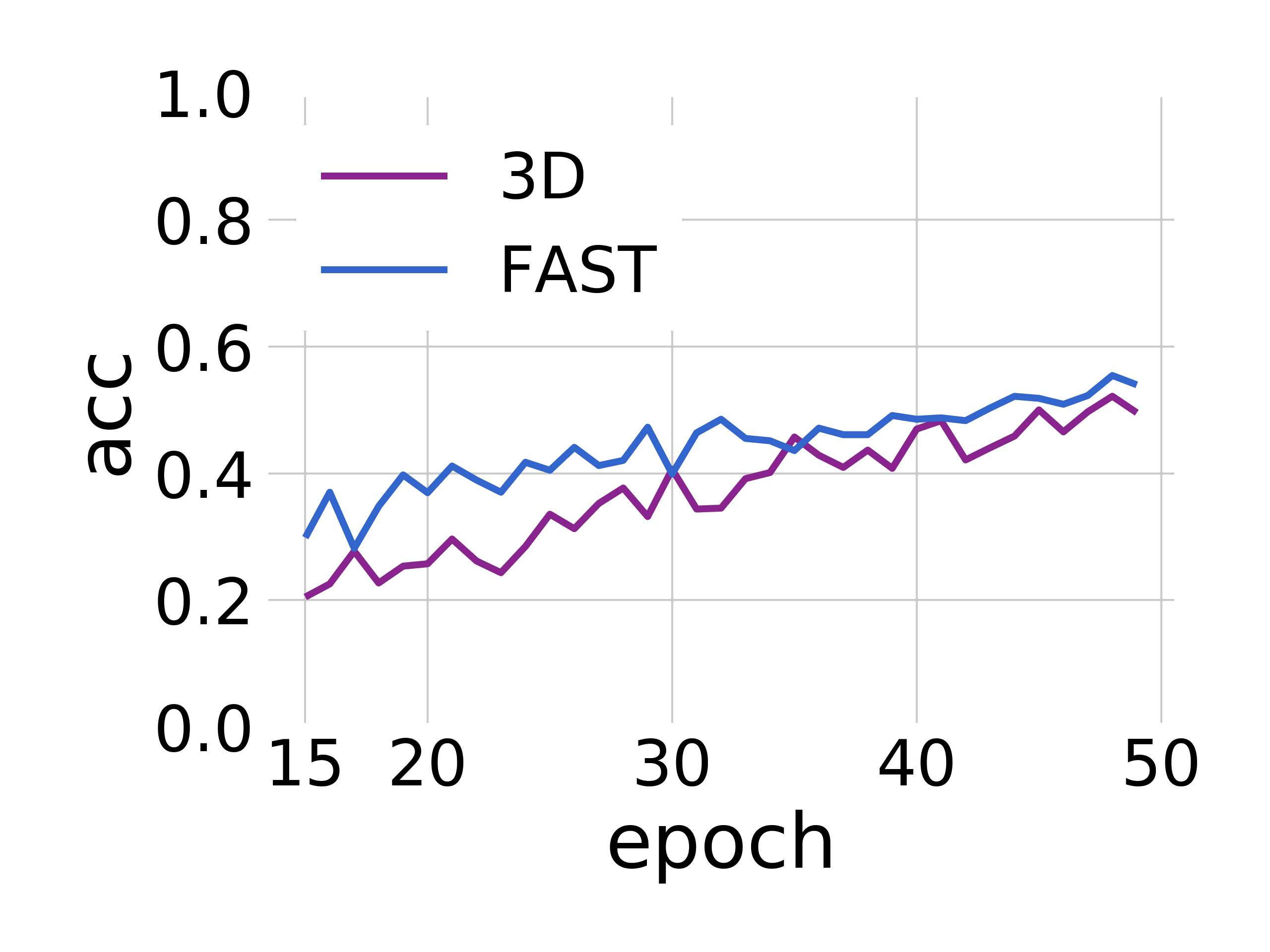}}
\end{center}
   \caption{\textbf{Validation loss and accuracy} for DenseNet-121 on UCF-101 (top row) and HMDB-51 (bottom row). The first 15 epochs are omitted.}
\label{fig:loss_acc_densenet}
\end{figure}

\subsection{Revisiting network architectures with 3D and FAST 3D convolutions}
\label{sec:Revisitingarchitectures}
We now investigate whether FAST 3D convolutions still perform better when the CNN architecture depth increases. To this end, we use ResNet-34, -50 and -101~\cite{he2016deep} and DenseNet-121~\cite{huang2017densely} architectures and replace the 2D convolution blocks either by 3D convolutions or FAST 3D convolutions. In addition to training and testing on UCF-101, we also use HMDB-51. The smaller size of this dataset in combination with the increasing depth of the network architectures allows us to investigate the risk of overfitting.

% results
Table~\ref{table:table3} summarizes the results of our tests. As expected, performance increases with the depth of the networks in general. In addition, FAST 3D convolutions consistently outperform 3D convolutions in every tested network and on both datasets. For networks with 34 and 50 layers, FAST 3D convolution blocks present an improvement of 0.2\% and 1.84\% respectively on UCF-101, and 0.78\% and 0.41\% on HMDB-51. For ResNet-101, the divergence between the two blocks increases to 2.62\% and 1.48\% for the two datasets, respectively. There is a direct correlation between the overall depth of the architecture and the performance difference between the two blocks. It appears that deeper layers of networks that focus on high level spatio-temporal features benefit from the separation into two spatio-temporal convolutions.

% densenet and comparison of loss
The densely connected and deeper DenseNet-121 architecture provides better results for both 3D and FAST 3D convolution blocks, on average 5.75\%-7.39\% better than ResNet-101. We also find that in this network, the improvements of FAST 3D over regular 3D convolutions are 0.6\% and 3.26\% for UCF-101 and HMDB-51, respectively. This difference is more modest but might be explained by the limited number of epochs. In Figure~\ref{fig:loss_acc_densenet}(a) and (c), it can be seen that the validation loss is still decreasing for both tested datasets. In Figure~\ref{fig:loss_acc_densenet}(b) and (d), it becomes clear that this also affects the validation accuracy, which might be even higher.
%This seems to be true for both tested convolution blocks.

\begin{table}[htb]
\begin{center}
\begin{tabular}{|l c c c c|}
\hline
\multirow{2}{*}{Method} & \multicolumn{2}{c}{UCF-101} & \multicolumn{2}{c|}{HMDB-51} \\
& 3D & FAST-3D & 3D & FAST-3D \\
\hline\hline
ResNet-34 & 81.14 & \textbf{83.82} & 39.68 & \textbf{40.16}\\
\hline
ResNet-50 & 81.68 & \textbf{84.62} & 45.44 & \textbf{45.95}\\
\hline
ResNet-101 & 82.17 & \textbf{84.79} & 46.53 & \textbf{48.01}\\
\hline
DenseNet-121 & 88.93 & \textbf{89.53} & 52.14 & \textbf{55.40}\\
\hline
\end{tabular}
\end{center}
\caption{Increasingly deep CNN architectures with either 3D or FAST 3D convolution blocks, trained and tested on either UCF-101 or HMDB-51. The proposed FAST 3D convolutions consistently outperform 3D convolutions in every tested architecture.}
\label{table:table3}
\end{table}

\subsection{Network comparisons}
\label{sec:Networkcomparisons}
To understand the overall performance of our introduced FAST 3D convolution block, we compare its performance to CNN architectures that are based on a separate treatment of spatial 2D convolutions and optical flow, and based on 3D convolutions. For the former category, we compare against the Two-stream approach \cite{simonyan2014two}, based on two VGG-16 networks for RGB and optical flow inputs, respectively. Long-Short-Term-Memory (LSTM) networks have been used for the fusion of the spatial and temporal information from the two streams \cite{yue2015beyond} (Two-stream + LSTM). Finally, we evaluate the performance of Temporal Segment Networks \cite{wang2016temporal}, where videos are divided into three segments with each of their predicted classes fused to obtain a final score. We also compare against several CNN architectures based on 3D convolutions, including the C3D network \cite{tran2015learning}. We further investigate different architectures of the proposed Pseudo 3D convolutions (P3D, \cite{qiu2017learning}) and (2+1)D Resnet-152 \cite{tran2018closer} that incorporate $1 \times 3 \times 3$ spatial kernels and $3 \times 1 \times 1$ temporal kernels. Lastly, we replicate a spatial I3D network \cite{carreira2017quo}. The memory requirements of other, more recent, networks with higher reported scores on both UCF-101 and HMDB-51 prevent us from testing these models.

% implementation
All models have been trained and tested on their datasets, and on the same machine. Differences between reported numbers in literature are largely due to the batch size, which we fixed to 8. This allows for a fair comparison. Typically, performance will go up once large batch sizes can be processed.

\begin{table}[htb]
\begin{center}
\begin{tabular}{|l c c |}
\hline
Method & UCF-101 & HMDB-51 \\
\hline\hline
\multicolumn{3}{|c|}{2D CNNs with Two-stream approach} \\
\hline\hline
Two-stream \cite{simonyan2014two} & 73.0 & 40.5  \\
\hline
Two-stream + LSTM \cite{yue2015beyond} & 82.6 & 47.1  \\
\hline
TSN \cite{wang2016temporal} & 85.7 & 54.6  \\
\hline\hline
\multicolumn{3}{|c|}{3D CNNs} \\
\hline\hline
C3D \cite{tran2014activity} & 44.9 & 43.9  \\
\hline
P3D \cite{qiu2017learning} & 83.2 & 45.1  \\
\hline
(2+1)D ResNet152 \cite{tran2018closer} & 85.7 & 45.8  \\
\hline
RGB-I3D \cite{carreira2017quo} & 86.4 & 53.2  \\
\hline\hline
\multicolumn{3}{|c|}{2D CNNs converted to FAST 3D convolutions} \\
\hline\hline
ResNet-50 & 84.6 & 45.9  \\
\hline
ResNet-101 & 84.7 & 48.0 \\
\hline
DenseNet-121 & 89.5 & 55.4  \\
\hline
\end{tabular}
\end{center}
\caption{Accuracy rates of CNN models trained on UCF-101 and HMDB-51 datasets, divided into architectures that use 2D spatial and temporal information in separate processes (Two-stream approach), 3D convolutions and the proposed FAST 3D convolutions are replacements of 2D convolutions. \label{table:table2}}
\end{table}

% Two-stream vs 3D
The performance of all tested architectures is summarized in Table~\ref{table:table2}. Several conclusions can be drawn from these results. First, the effective modeling of temporal characteristics is important. There is clear performance gain of LSTM fusion and TSN over the regular Two-stream results. This is primarily because the Two-stream approach is limited by processing the spatial features in a per-frame fashion, and only considers temporal information between subsequent frames. In contrast, 3D convolutions are trained over small spatio-temporal slices and thus consider the temporal nature to a larger extent. 
%This was also part of our motivation for using 24 frame clips for our experiments as a smaller number of frames would not demonstrate this difference.

% Results based on architectures
DenseNet-121 with FAST 3D convolutions is the best performing tested architecture for both UCF-101 and HMDB-51. Due to the larger number of skip connections in DenseNet, in every pass both low-level and high level spatio-temporal features are learned. It is clear that the decoupling of the 3D spatio-temporal input into orthogonal spatio-temporal inputs benefits from this. It is expected that the use of split-FAST could further increase the performance.

%-------------------------------------------------------------------------
\section{Conclusion}
\label{Conclusion}
% summary
We have introduced FAST 3D convolutions, a novel convolution block that combines a 2D spatial convolution with two orthogonal spatio-temporal convolutions. The block is motivated by the often characteristic horizontal and vertical motion of human actions. In experiments on UCF-101 and HMDB-51, the novel FAST 3D convolutions consistently outperform 3D convolutions on ResNets with several depths and DenseNet-121. We also presented a split-FAST block with both motion directions in separate pathways, which increased performance even further. The novel blocks generalize somewhat better, based on the lower validation loss. In a comparison with CNN architectures with similar memory requirements, DenseNet-121 with FAST 3D convolutions scored best. 

% limitations and future work
Future experiments should additionally consider recently introduced large action recognition datasets such as Kinetics \cite{kay2017kinetics} and ActivityNet \cite{heilbron2015activitynet}. The FAST 3D convolution block can be used in many CNN architectures. Adoption of the block in state-of-the-art network architectures such as Squeeze-and-Excitation Networks~\cite{hu2018squeeze} and Neural Architecture Search Networks~\cite{Zoph_2018_CVPR} appears a promising direction to address human action recognition tasks.

\section{Acknowledgments}
\label{acknowledgments} \label{sec:section7}
This work is supported by the Netherlands Organization for Scientific Research (NWO) with a TOP-C2 grant for “Automatic recognition of bodily interactions” (ARBITER).

%------------------------------------------------------------------------

{\small
\bibliographystyle{ieee}
\bibliography{main}
}

\end{document}